\documentclass[letterpaper, 10 pt, conference]{ieeeconf}  
\IEEEoverridecommandlockouts                              

\usepackage{graphicx} 
\usepackage{times} 
\usepackage{cite}
\usepackage{amsmath} 
\usepackage{amssymb}  
\usepackage{braket}
\usepackage{booktabs}
\usepackage{colortbl}
\usepackage{url}
\usepackage{mathtools}
\usepackage[linesnumbered,ruled,vlined]{algorithm2e}

\usepackage[caption=false, font=footnotesize]{subfig}

\usepackage{lipsum}

\newcommand{\latent}{{\boldsymbol z}}
\newcommand{\obs}{{\boldsymbol x}}
\newcommand{\predm}{{\boldsymbol \mu}}
\newcommand{\preds}{{\boldsymbol \sigma}}
\newcommand{\gt}{{\boldsymbol \mu_{i}^{*}}}
\newcommand{\mota}[1]{\begin{tabular*}{1.1cm}{c}MOTA \\ #1 \end{tabular*}}
\newcommand{\bhline}[1]{\noalign{\hrule height #1}}

\title{\LARGE \textbf{
Multi-person Pose Tracking using Sequential Monte Carlo \\ with Probabilistic Neural Pose Predictor
}}

\author{Masashi Okada$^{\dag,\star}$, Shinji Takenaka$^{\ddag}$ and Tadahiro Taniguchi$^{\dag,*}$
    \thanks{
        $^{\dag}$ Masashi Okada and Tadahiro Taniguchi are with AI Solutions Center, Business Innovation Division, Panasonic Corporation, Japan.
    }%
    \thanks{
        $^{\ddag}$ Shinji Takenaka is with Technology Center, Panasonic System Networks R\&D Lab.~Co.,~Ltd., Japan.
    }%
    \thanks{
        $^{*}$ Tadahiro Taniguchi is also with Ritsumeikan University, College of Information Science and Engineering, Japan.
    }%
    \thanks{
        $^{\star}$ Corresponding author: \texttt{okada.masashi001@jp.panasonic.com}
    }
}

\begin{document}

\maketitle
\thispagestyle{empty}
\pagestyle{empty}

\begin{abstract}
It is an effective strategy for the multi-person pose tracking task in videos to employ prediction and pose matching in a frame-by-frame manner.
For this type of approach, uncertainty-aware modeling is essential because precise prediction is impossible.
However, previous studies have relied on only a single prediction without incorporating uncertainty, which can cause critical tracking errors if the prediction is unreliable.
This paper proposes an extension to this approach with Sequential Monte Carlo (SMC).
This naturally reformulates the tracking scheme to handle multiple predictions (or hypotheses) of poses, thereby mitigating the negative effect of prediction errors.
An important component of SMC, i.e.,~a proposal distribution, is designed as a \textit{probabilistic neural pose predictor},
which can propose diverse and plausible hypotheses by incorporating epistemic uncertainty and heteroscedastic aleatoric uncertainty.
In addition, a recurrent architecture is introduced to our neural modeling to utilize time-sequence information of poses to manage difficult situations,
such as the frequent disappearance and reappearances of poses.
Compared to existing baselines, the proposed method achieves a state-of-the-art MOTA score on the PoseTrack2018 validation dataset by reducing approximately 50\% of tracking errors from a state-of-the art baseline method.

\end{abstract}

\section{Introduction}
Object detection and tracking are important tasks for various robotics applications, such as
autonomous  vehicle control~\cite{behrendt2017deep,verma2018vehicle,buyval2018realtime,sharma2018beyond},
visual SLAM~\cite{concha2015dpptam,liang2018planar}, and
robotics manipulator control~\cite{rauch2018visual}.
Multi-person pose estimation and tracking is a critical component in various applications, such as video surveillance and sports video analytics.
In the past few years, pose estimation has progressed significantly~\cite{cao2017realtime} due to deep convolutional learning assisted by large-scale image corpora, such as COCO~\cite{lin2014microsoft} and MPPI~\cite{andriluka14cvpr}.
The PoseTrack dataset~\cite{iqbal2017posetrack} is a video corpus for pose estimation and tracking that is annotated with multiple people in scenes
and this dataset has encouraged the community to develop a diverse range of pose estimation and tracking models~\cite{girdhar2018detect,xiu2018pose,guo2018multi,xiao2018simple,hwang2019pose,raaj2019efficient}.

Most of these pose tracking models employ a two-stage scheme, i.e.,~\textit{1)} poses are estimated using a deep convolutional neural network (CNN), and then \textit{2)} poses are tracked by employing greedy bipartite matching in a frame-by-frame manner.
For example, \textit{Simple-Baseline}~\cite{xiao2018simple} (the Pose Track Challenge ECCV'18 Winner) introduces matching utilizing \textit{flow-based pose similarity},
which is defined as the \textit{object keypoint similarity} (OKS)~\cite{lin2014microsoft} between a pose estimated from a current frame and a pose predicted from previous results using optical-flows.
This optical-flow-based prediction can compensate the pose differences of the same person across multiple frames, which making the matching scores robust against fast movements of people and cameras.
We believe that this type of prediction-based matching is general and strong approach,
and can be applied to not only for pose tracking but also various multi-object tracking tasks.

\begin{figure}[t]
  \centering
  \includegraphics[width=0.475\textwidth]{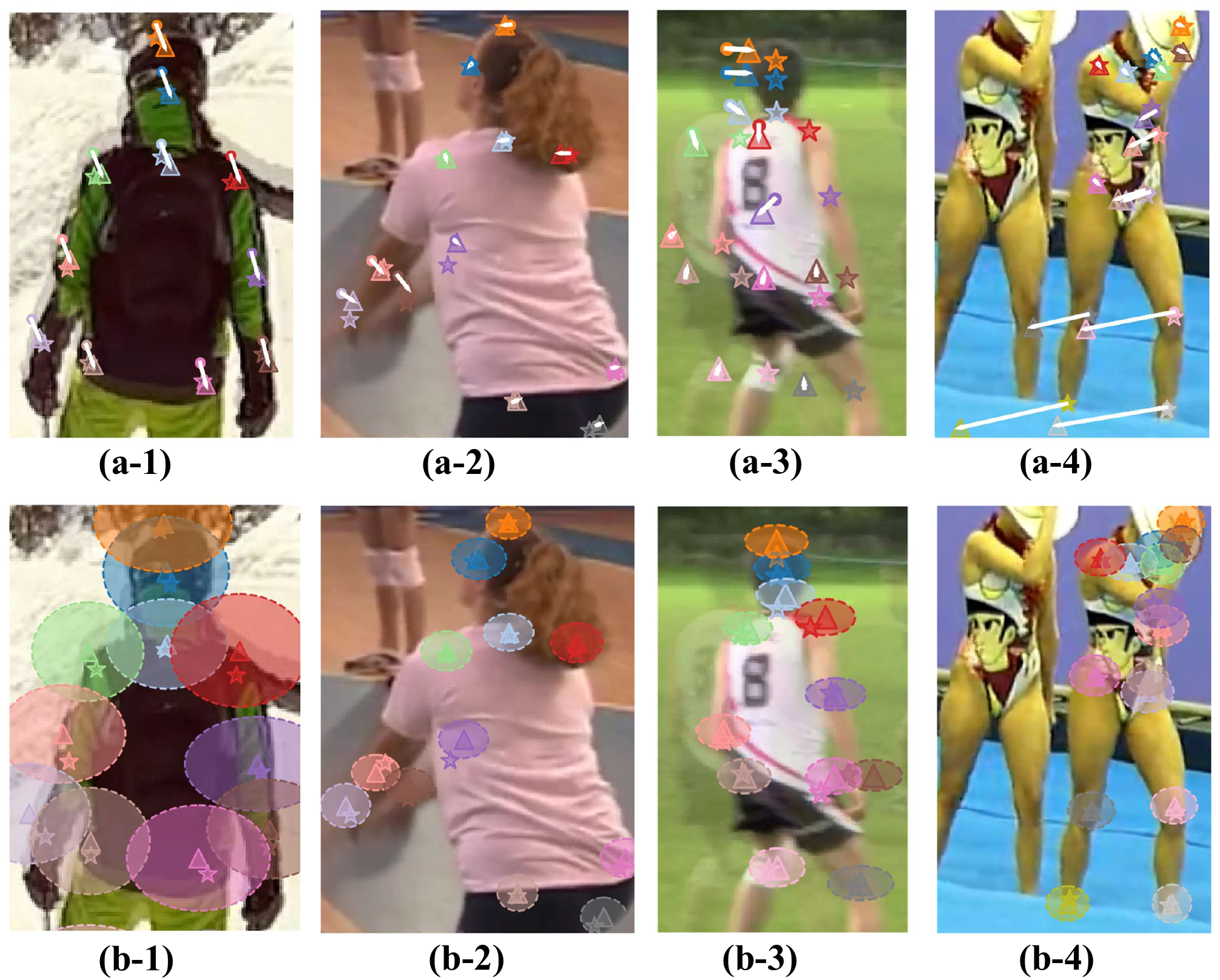}
  \caption{
  Human poses predicted by (a) optical-flows and (b) the proposed probabilistic neural predictor.
  Two consecutive frames are rendered with $\alpha$-blending with ground-truth current keypoints shown as `$\star$'s.
  In (a), `$\bullet$'s and `$\triangle$'s indicate previous and predicted keypoints, respectively.
  The solid lines between `$\bullet$'s and `$\triangle$'s illustrate optical-flows obtained by a state-of-the-art
  optical-flow estimator PWC-Net~\cite{sun2018pwc}.
  In (b), `$\triangle$'s and ovals with dashed-lines respectively indicate the average and deviation ($2\sigma$) of 100 different particles
  predicted by inputting the same 100 inputs to the predictor.
  }
  \label{fig:predictions}
\end{figure}
\begin{figure*}[t]
  \centering
  \includegraphics[width=0.75\textwidth]{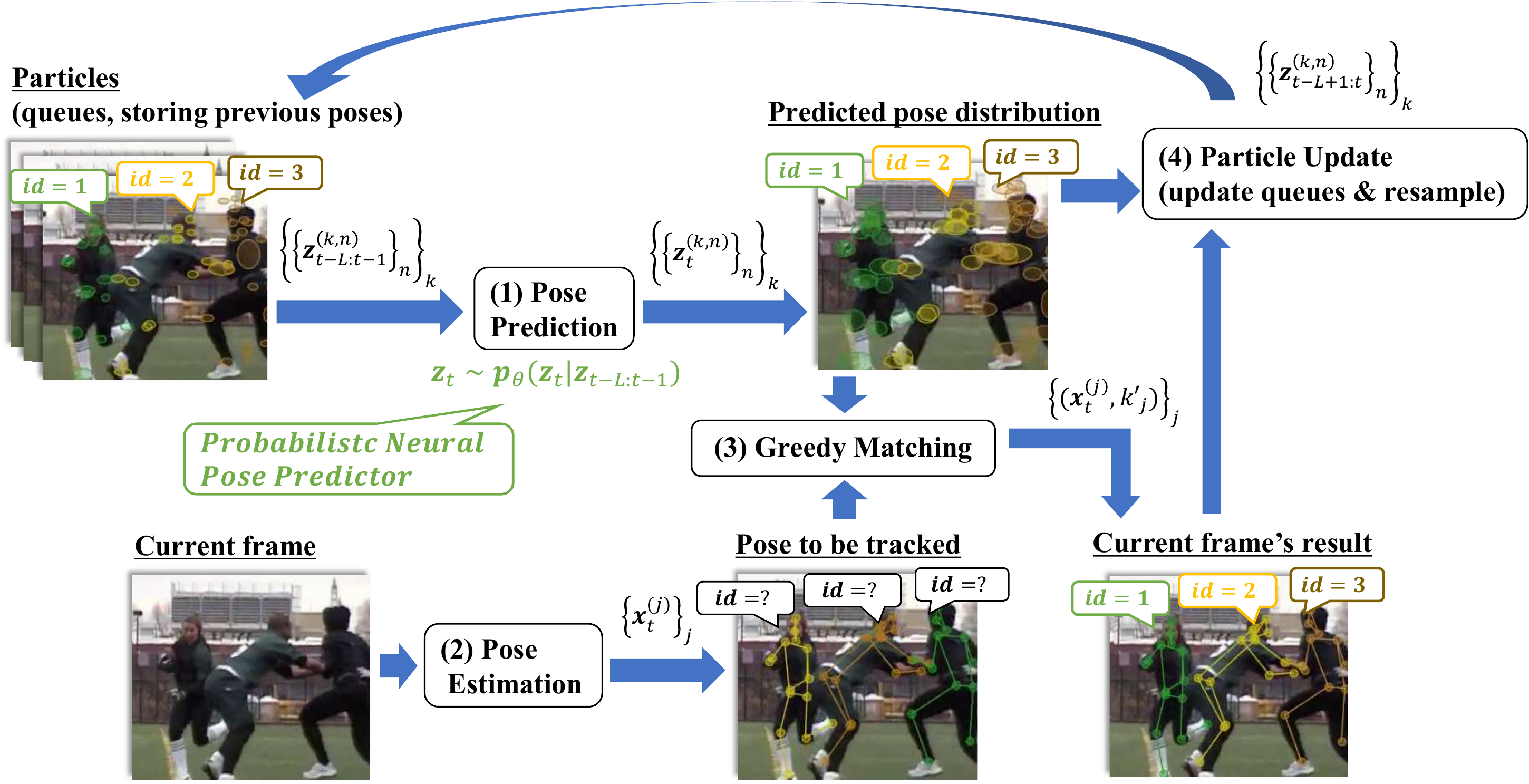}
  \caption{Proposed SMC-based pose tracking method.
  The details of this method will be described in Sec.~\ref{sec:preliminary}--\ref{sec:proposed_tracking_method}.}
  \label{fig:proposed_method}
\end{figure*}

However, the matching process strongly relies on a single hypothesis, which could be critically vulnerable to prediction errors.
The errors cause underestimation of the matching score between correct pairs, thereby resulting in mismatching.
An optical-flow based prediction can give reasonable predictions, as demonstrated in Fig.~\ref{fig:predictions}~(a-1);
however, this can also generate unreliable predictions especially when
poses change quickly or there is insufficient texture information available around a given keypoint (see Figs.~\ref{fig:predictions}~(a-2)--(a-4) for examples).
A possible solution to this is pursuing prediction accuracy; however, precise prediction appears impossible due to
uncertainties~\cite{kendall2017uncertainties} in human gait, imaging, pose estimation by deep CNN models, etc.
Another concern is that optical-flows can be sensitive to the disappearance and reappearance of poses caused by occlusions.
The baseline introduces multiple optical-flows of previous multi-frames to mitigate this issue;
however, estimating accurate multi-frame optical-flows requires complicated processes~\cite{janai2018unsupervised,ren2019fusion}.

Motivated by this, we have designed an extenstion to this tracking strategy.
Note that this paper primarily focuses on the pose tracking methodology
and does not discuss about CNN modeling for pose estimation.
In the tracking system we implemented, the CNN pose estimator from the \textit{Simple-Baseline} is adopted without modification.
Our primary contributions are summarized as follows.
\begin{itemize}
  \item We propose a multi-person tracking method that exploits the newly-devised \textit{probabilistic neural pose predictor}
  and the well-known Sequential Monte Carlo (SMC; or \textit{particle filter}), the diagram of which is illustrated in Fig.~\ref{fig:proposed_method}.
  \item The proposed tracking method achieves a state-of-the-art Multiple Object Tracking Accuracy (MOTA)~\cite{bernardin2006multiple} score of 66.2 on the PoseTrack2018 validation dataset. Our method outperforms both the ECCV'18 Winner \textit{Simple-Baseline}~\cite{xiao2018simple} (MOTA:~65.4) and a more recent baseline method~\cite{hwang2019pose} (MOTA:~65.7).
\end{itemize}
Our probabilistic neural pose predictor, as a principal component of SMC,
probabilistically predicts the next poses considering two kinds of
epistemic uncertainty (model uncertainty due to limited data) and heteroscedastic aleatoric uncertainty (inherent system stochasticity).
This stochasticity of the predictor allows us to prepare diverse predictions (or hypotheses).
To consider long context information, a recurrent neural modeling with Long Short-Term Memory (LSTM)~\cite{hochreiter1997long} is introduced
to manage difficult situations, such as frequent occlusions.
Figure~\ref{fig:predictions}~(b) illustrates how the predictor infers plausible pose distributions.

The remainder of this paper is organized as follows.
In Sec.~\ref{sec:preliminary}, we briefly review SMC and formulate a single pose tracking problem with SMC.
In Sec.~\ref{sec:predictor}, we summarize the concept and architecture of the pose predictor.
%
Sec.~\ref{sec:proposed_tracking_method} proposes a tracking method that exploits this predictor.
%
In Sec.~\ref{sec:experiments}, the effectiveness of the proposed method is demonstrated in an evaluation using the PoseTrack2018 dataset.

\section{SMC for Single Object Tracking} \label{sec:preliminary}
\begin{figure}
  \centering
  \includegraphics[width=0.2\textwidth]{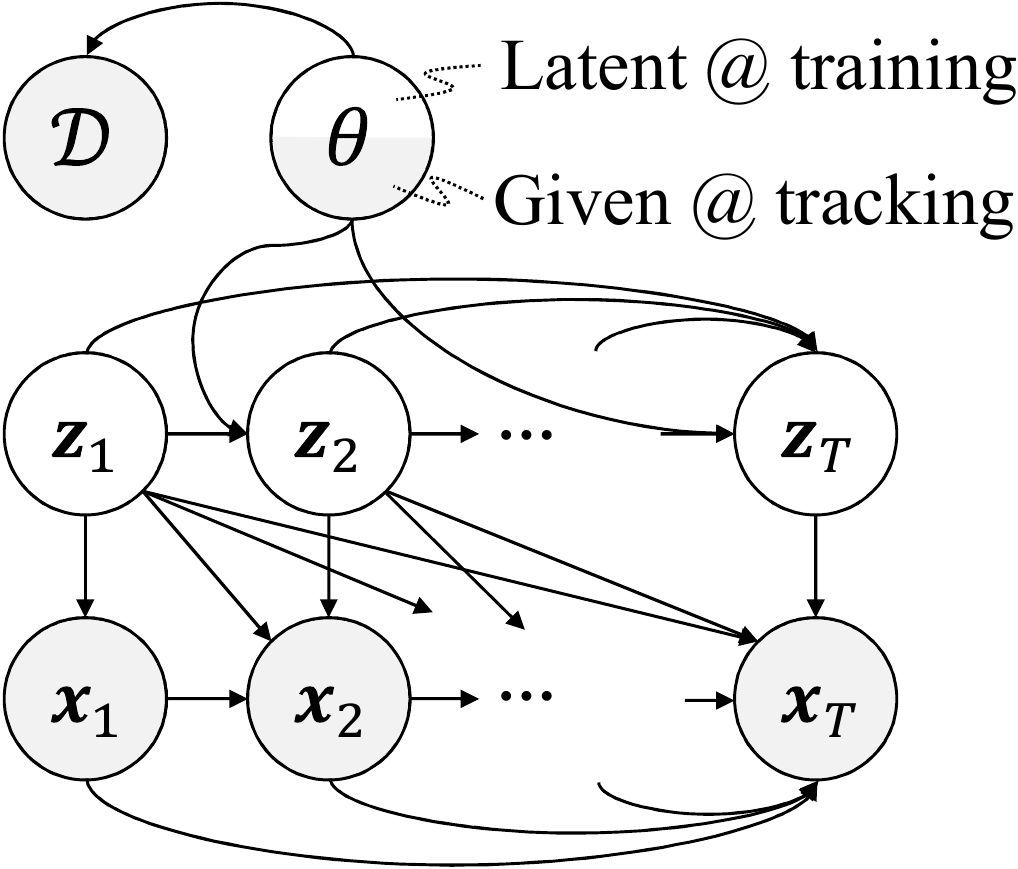}
  \caption{Graphical model of pose track formulation.}
  \label{fig:graphical_model}
\end{figure}

This section briefly describes an SMC formulation for a single object tracking task.
Fig.~\ref{fig:graphical_model} shows the graphical model for this formulation,
which takes a more general form subsuming common state-space models, such as Hidden Markov Models.
This model comprises latent states $\latent_{1:T}$ and observed states $\obs_{1:T}$.
Latent state $\latent_{t}$ can be predicted by a transition model $\latent_{t} \sim p(\latent_{t}|\latent_{1:t-1}, \theta)$ parameterized with $\theta$,
the posterior $p(\theta|\mathcal{D})$ of which is inferred from given training dataset $\mathcal{D} = \{(\latent_{1:t-1}, \latent_{t})\}$.
The joint distribution of this model takes the factorized form
$p(\latent_{1:T}, \obs_{1:T}, \theta) = p(\latent_{1})p(\obs_{1}|\latent_{1})\prod^{T}_{t=2}p_{\theta}(\latent_{t}|\latent_{1:t-1})p(\obs_{t}|\latent_{1:t}, \obs_{1:t-1})$,
where $p_{\theta}(\latent_{t}|\latent_{1:t-1}) \coloneqq p(\latent_{t}|\latent_{1:t-1}, \theta)p(\theta|\mathcal{D})$.
The objective of SMC is to approximately infer the posterior over the latent state sequence with a set of $N$ \textit{particles} as
\begin{equation}
  p(\latent_{1:T}|\obs_{1:T}) \simeq \sum^{N}_{n=1}w(\latent^{(n)}_{1:T})\delta(\latent_{1:T}-\latent^{(n)}_{1:T}),
\end{equation}
where $\delta$ is the Dirac delta function.
%
%
%
%
%
%
$w(\latent^{(n)}_{1:T})$ is the weight of a particle $n$, which can be recursively defined as
\begin{equation}
  w(\latent^{(n)}_{1:T})
  \propto
  w(\latent^{(n)}_{1:T-1}) \frac
  {p_{\theta}(\latent^{(n)}_{T}|\latent^{(n)}_{1:T-1})p(\obs_{T}|\latent^{(n)}_{1:T},\obs_{1:T-1})}
  {q(\latent^{(n)}_{T}|\latent^{(n)}_{1:T-1},\obs_{1:T})}, \label{eqn:weight_update_law}
\end{equation}
where $q(\latent^{(n)}_{T}|\latent^{(n)}_{1:T-1},\obs_{1:T})$ is a proposal distribution for importance sampling.
Note that this proposal distribution must be defined carefully to propose plausible particles.
A similar derivation of \eqref{eqn:weight_update_law} can be found in the literature~\cite{gu2015neural}.

In this formulation, particles are managed according to the following procedures.
First, at time step $T$, new particles $\latent^{(n)}_{T}$ are proposed from $q(\latent^{(n)}_{T}|\latent^{(n)}_{1:T-1},\obs_{1:T})$.
Second, the particles are weighted by \eqref{eqn:weight_update_law},
and then particles $\latent^{(n)}_{1:T}$ are \textit{resampled} from a categorical distribution $\operatorname{Cat}$ as
\begin{equation}
  \latent^{(n)}_{1:T} \sim \operatorname{Cat}\left(N, \{w(\latent^{(n)}_{1:T})\}_{n}\right). \label{eqn:cat}
\end{equation}
After resampling, all weights $w(\latent^{(n)}_{1:T})$ are made uniform.

In our case, $\obs_{t}$ and $\latent^{(n)}_{t}$ correspond to the estimated and predicted poses of a single person, respectively.
Managing multiple particles allows us to have multiple pose hypotheses.
The likelihood (or confidence) of each hypothesis is evaluated by the second term in the nominator of \eqref{eqn:weight_update_law}.
This term is referred to as a \textit{likelihood function}, and we define this function with OKS between $\obs_{T}$ and $\latent^{(n)}_{T}$ as
\begin{equation}
  p(\obs_{T}|\latent^{(n)}_{1:T},\obs_{1:T-1}) \propto \operatorname{OKS}(\obs_{T}, \latent^{(n)}_{T}). \label{eqn:oks}
\end{equation}
We assume the proposal distribution takes the form
\begin{align}
  & q(\latent^{(n)}_{T}|\latent^{(n)}_{1:T-1},\obs_{1:T}) \nonumber \\ & \coloneqq p_{\theta}(\latent^{(n)}_{T}|\latent_{1:T-1})q(\latent_{1:T-1}|\latent^{(n)}_{1:T-1}, \obs_{1:T-1}). \label{eqn:proposer1}
\end{align}
The second term on the right-hand side is defined as
\begin{align}
  & {q(\latent_{1:T-1}|\latent^{(n)}_{1:T-1}, \obs_{1:T-1})} \nonumber = \\ &  \alpha \delta(\latent_{1:T-1} - \latent^{(n)}_{1:T-1}) + (1 - \alpha) \delta(\latent_{1:T-1} - \obs_{1:T-1}), \label{eqn:proposer2}
\end{align}
which selects $\latent^{(n)}_{1:T-1}$ or $\obs_{1:T-1}$ probabilistically as the input to prediction model $p_{\theta}(\latent^{(n)}_{T}|\latent_{1:T-1})$.
The parameter $\alpha \in [0, 1]$ controls how we weight the prediction and observation.
We found that $\alpha = 0.45$ achieves the best performance on multi-person pose track formulation.
In summary, \eqref{eqn:weight_update_law} can be described as
\begin{equation}
  w(\latent^{(n)}_{1:T})
  \propto
  w(\latent^{(n)}_{1:T-1}) \frac
  {\operatorname{OKS}(\obs_{T}, \latent^{(n)}_{T})}
  {q(\latent_{1:T-1}|\latent^{(n)}_{1:T-1}, \obs_{1:T-1})}. \label{eqn:weight_update_law2}
\end{equation}

In Sec.~\ref{sec:predictor}, we discuss how we design predictor $p_{\theta}(\cdot)$ to propose plausible hypotheses,
and, in Sec.~\ref{sec:proposed_tracking_method}, we extend this SMC scheme to multi-person pose tracking.

\section{Probabilistic Neural Pose Predictor} \label{sec:predictor}
\begin{figure*}
  \centering
  \includegraphics[width=0.75\textwidth]{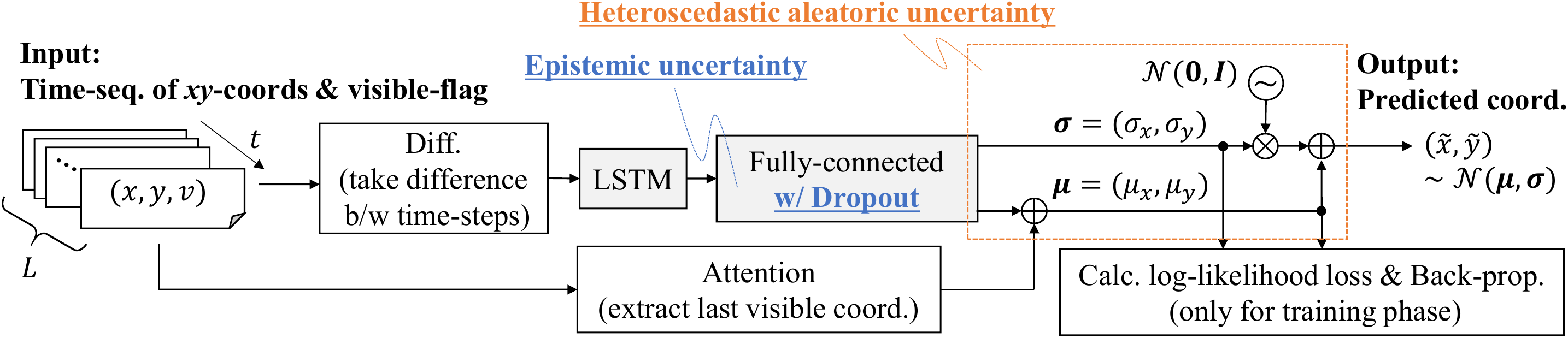}
  \caption{
  Architecture of probabilistic neural pose predictor.
  Prediction is performed in a pointwise manner, i.e.,~each keypoint of a pose is processed independently.
  Shaded boxes indicate trainable modules parameterized with $\theta$,
  which are trained to estimate the residual between the last visible coordinates and the current prediction.
  The number of LSTM units is 64.
  The fully connected module has one hidden layer with 40 hidden nodes and the leaky-ReLU activation function.
  The dropout probability of the hidden nodes is 0.3.
}
  \label{fig:predictor}
\end{figure*}
We model predictor $p_{\theta}(\cdot)$ as a trainable neural network.
Fig.~\ref{fig:predictor} illustrates the architecture of this model, which we refer to as the \textit{probabilistic neural pose predictor}.
This predictor is designed to have probabilistic behaviors by incorporating
epistemic uncertainty and heteroscedastic aleatoric uncertainty~\cite{kendall2017uncertainties},
which allow us to propose multiple hypotheses satisfying both diversity and plausibility.
In addition, to utilize the time-sequence input $z_{1:T-1}$, we employ recurrent neural modeling by LSTM~\cite{hochreiter1997long}
with a stateless architecture.
Here, the time length is constrained to $L$, thus $z_{T-L:T-1}$ is input to the model.

\subsection{Epistemic Uncertainty}
Epistemic uncertainty accounts for uncertainty in the model parameters $\theta$ due to a lack of sufficient data $\mathcal{D}$,
i.e.,~$p(\theta|\mathcal{D})$, which is also referred to as model uncertainty.
If nearly infinite data is available, this uncertainty should vanish.
However, in a practical case where $\mathcal{D}$ is insufficient and/or a model has a deep (or overparameterized) architecture,
this uncertainty remains and should be managed carefully.
We exploit \textit{dropout as inference}~\cite{gal2016dropout} to model this uncertainty,
which approximates $p(\theta|\mathcal{D})$ as a Gaussian distribution $q(\theta)$.
It has been proven that the variational inference problem, i.e.,~$\operatorname{argmin}_{q}\operatorname{KL}(q(\theta)||p(\theta|\mathcal{D}))$, is approximately equivalent to training networks with dropout, where $\operatorname{KL}(\cdot||\cdot)$ denotes Kullback-Leibler divergence.
In our modeling, dropout is performed both in the training and test steps.
Consequently, even if the same inputs are fed into the predictor during the test step, different models are sampled probabilistically from $q(\theta)$ and applied to each input, thereby proposing diverse hypotheses.
%

\subsection{Heteroscedastic Aleatoric Uncertainty}
Aleatoric uncertainty represents noise inherent in observations.
In pose prediction, this uncertainty can originate from sudden changes of human gaits,
fast camera panning and tilting, and pose estimation errors by deep CNN models.
We model aleatoric uncertainty with heteroscedasticity.
Specifically, our predictor is trained to estimate input-dependent Gaussian distributions and
adaptively change the diversity (or variance) of particles according to different situations.
For example, in cases in which people move very quickly, the predictor scatters particles in a wider area.
In cases with slower movement, the predictor concentrates particles in a narrower area.
This behavior helps us utilize the finite particles effectively.

In the training step, model parameter $\theta$ is optimized to minimize the log-likelihood loss defined as
\begin{align}
  &-\log p(\theta|\mathcal{D}) \propto - \log p(\mathcal{D}|\theta)p(\theta)
  \propto \nonumber\\
  & \sum_{i}\left\{(\predm_{i} - \gt)^{T}\preds^{-1}(\predm_{i} - \gt) + \log|\preds_{i}|\right\} + \lambda ||\theta||^{2},
  \label{eqn:loglikelihood}
\end{align}
where $i$ is the index of an element in dataset $\mathcal{D}$.
$\predm_{i}$, $\preds_{i}$ and $\predm^{*}_{i}$ respectively denote estimated mean, deviation and ground-truth for input $i$.
The last term in (\ref{eqn:loglikelihood}) is an L2 regularization term originating from the Gaussian posterior $p(\theta)$.
In the test step, a single point is sampled from the estimated Gaussian distribution.

\section{Proposed Tracking Method} \label{sec:proposed_tracking_method}
Algorithm~\ref{alg:pose_track} gives the pseudocode for the proposed tracking method.
We implemented this method using TensorFlow~\cite{tensorflow2015-whitepaper}, and
all independent threads and for-loops except for the outermost loop ($\ell$\ref{alg:outermost}) can be executed efficiently in parallel on GPUs.
Our prototype implementation can simultaneously track $10$ poses at about $30$~fps on a single NVIDIA RTX2080 GPU. %
%
The rest of this section explains the variables and procedures of Algorithm.~\ref{alg:pose_track}.

\subsection{Notations and Internal Variables}
The number of estimated poses from current frame $t$ and their (tentative) index are denoted as $C_{t}$ and $j$, respectively.
The tracking system manages at most $F_{max}$ \textit{filters} to track multi-person poses,
each of which has a unique track ID $k$ and handles $P$ particles.
Particle $n$ of filter $k$ comprises an $L$-sized queue that stores previous poses $\latent^{(k,n)}_{t-L:t-1}$.
During the tracking process at $t$, only \textit{active} filters are executed.
The activation and deactivation of filters are controlled by \textit{lifetime counts} $l_{k}$ which manages the appearance and disappearance of people.
A set of active filters at $t$ are denoted as $\mathcal{K}_{t}$ ($|\mathcal{K}_{t}| \coloneqq F_{t} \leq F_{max}$).
Note that all of the filters are inactive when the algorithm just starts ($\ell$\ref{alg:start}).
The experiments in Sec.~\ref{sec:experiments} used $P=300$, $L=10$, and $F_{max}=100$.

\subsection{Module Procedures}

\renewcommand{\theparagraphdis}{\arabic{paragraph})}
\paragraph{Pose Prediction}
($\ell$\ref{alg:get_states}) $F_{t} \times P$ sequences of poses $\latent^{(k,n)}_{t-L:t-1}$ are collected from the active filters and then ($\ell$\ref{alg:pred_pose}) input to the predictor to output $F_{t} \times P$ predicted poses $\latent^{(k,n)}_{t}$.
Note that these procedures are not executed in case where there are no active filters
(e.g., $t=1$).

\paragraph{Pose Estimation}
($\ell$\ref{alg:estimate_pose}) $C_{t}$ poses $\obs^{(j)}_{t}$ are estimated by inputting a current RGB image to the deep CNN model of \textit{Simple-Baseline}~\cite{xiao2018simple}.
Here we used an opensource codebase%
\footnote{\url{https://github.com/mks0601/TF-SimpleHumanPose}}
to realize the CNN model and completely followed the experimental settings described in the literature~\cite{xiao2018simple} to train the model.

\paragraph{Greedy Matching}
($\ell$\ref{alg:calc_oks}) In total, $C_{t} \times F_{t} \times P$ OKS values $d^{(j,k,n)}_{OKS}$ are calculated from $F_{t} \times P$ predicted poses $\latent^{(k,n)}_{t}$ and $C_{t}$ estimated poses $\obs^{(j)}_{t}$.
($\ell$\ref{alg:calc_score}) The shape of this OKS tensor is transformed to $C_{t} \times F_{t}$ by taking the weight average along the $n$-axis to calculate a matching score matrix%
\footnote{
We determine particle weights as followings:
the weights of the top $e$\% particles with higher OKS are set to 1; the remaining weights are set to 0.
The \textit{eliteness} ratio $e$ is 15\%.
}.
Then, ($\ell$\ref{alg:match}) this matrix is input to the bipartite matching procedures.
Formed pairs whose matching score is less than a given threshold (i.e., two poses are distant to each other) are removed to prevent inappropriate matching.
Here, $j'_{k}$ and $k'_{j}$ denote the indices of counterparts of filter $k$ and pose $j$, respectively.
Note that these variables take negative values when no counterparts are assigned due to shortage and overage of the active filters, and the thresholding.
($\ell$\ref{alg:output0}--\ref{alg:output1}) If $k'_{j}$ has a valid value for pose $j$, a tuple of $({\obs^{(j)}_{t}}, k'_{j})$ is output as the tracking result.
($\ell$\ref{alg:else}) If $k'_{j}$ has an invalid value, (\ref{alg:activate}) a new filter $k_{new}$ is activated and (\ref{alg:output2}) this new index is output with $\obs^{(j)}_{t}$.
($\ell$\ref{alg:init1}--\ref{alg:init2}) The newly activated queue states are initialized with $\obs^{(j)}_{t}$ and zeros as invisible keypoints.

\paragraph{Particle Updates}
($\ell$\ref{alg:push}) The newest states $\latent^{(k,n)}_{t}$ are pushed to the queues (thereby removing the oldest ones).
($\ell$\ref{alg:update}--\ref{alg:resample}) If $j'_{k}$ has a valid value, the queues of filter $k$ are updated by
probabilistic resampling and selection according to \eqref{eqn:cat} and \eqref{eqn:proposer2}, respectively.
This process is not executed when $j'_{k} < 0$ because no information is available to update the confidences of the hypotheses.
($\ell$\ref{alg:increment}, $\ell$\ref{alg:decrement}) The lifetime count $l_{k}$ is incremented or decremented according the existence of filter $k$'s counterpart.
($\ell$\ref{alg:deactivate1}--\ref{alg:deactivate2}) If $\ell_{k}$ obtains zeros, a person tracked by filter $k$ tracks is considered to have disappeared completely; thus, the filter is deactivated.
%
\newcommand\mycommfont[1]{\textbf{\footnotesize{\color{blue}{#1}}}}
\SetCommentSty{mycommfont}
\SetAlCapNameFnt{\footnotesize}
\SetAlCapFnt{\footnotesize}

\begin{algorithm}
\footnotesize
\DontPrintSemicolon
  \textsc{DeactivateAllFilters()} \label{alg:start}\\
  \For{$t \leftarrow 1$ \KwTo $\infty$}{ \label{alg:outermost}
    \tcp*[l]{(1) Pose Prediction}
      $\{\{\latent^{(k,n)}_{t-L:t-1}\}^{P}_{n=1}\}_{k \in \mathcal{K}_{t}} \leftarrow $ \textsc{GetActiveFilterStates()} \label{alg:get_states}\\
      $\{\{\latent^{(k,n)}_{t}\}_{n}\}_{k} \leftarrow $ \textsc{PredictPose}$(\{\{\latent^{(k,n)}_{t-L:t-1}\}_{n}\}_{k})$ \label{alg:pred_pose}\\
    \tcp*[l]{(2) Pose Estimation}
    $\{\obs^{(j)}_{t}\}^{C_{t}}_{j=1} \leftarrow $ \textsc{EstimatePosesFromCurrentFrame()} \label{alg:estimate_pose}\\
    \tcp*[l]{(3) Greedy Matching}
    $\{\{\{d^{(j,k,n)}_{OKS}\}_{j}\}_{k}\}_{n} \leftarrow$ \textsc{CalcOKS$\left(\{\{\latent^{(k,n)}_{t}\}_{n}\}_{k}, \{\obs^{(j)}_{t}\}_{j}\right)$} \label{alg:calc_oks} \\
    $\{\{d^{(j,k)}_{score}\}_{j}\}_{k} \leftarrow$ \textsc{CalcScore$\left(\{\{\{d^{(j,k,n)}_{OKS}\}_{j}\}_{k}\}_{n}\right)$} \label{alg:calc_score}\\
    $(\{j'_{k}\}_{k}, \{k'_{j}\}_{j}) \leftarrow$ \textsc{BipartiteMatch$\left(\{\{d^{(j,k)}_{score}\}_{j}\}_{k}\right)$} \label{alg:match}\\
    \For{$j \leftarrow 1$ \KwTo $C_{t}$}{
      \If{$k'_{j} > -1$}{\label{alg:output0}
        \textsc{Output$\left(\obs^{(j)}_{t}, k'_{j}\right)$} \label{alg:output1}\\
      }
      \Else{ \label{alg:else}
        $k_{new} \leftarrow$ \textsc{ActivateNewFilter()} \label{alg:activate}\\
        $l_{k} \leftarrow 1$ \label{alg:init0}\\
        $\{\latent^{(k_{new}, n)}_{t}\}_{n} \leftarrow \{\obs^{(j)}_{t}\}_{n}$ \label{alg:init1}\\
        $\{\latent^{(k_{new}, n)}_{t-L+1:t-1}\}_{n} \leftarrow \mathbf{0}$ \label{alg:init2}\\
        \textsc{Output$\left(\obs^{(j)}_{t}, k_{new}\right)$} \label{alg:output2}\\
      }
    }
    \tcp*[l]{(4) Particle Update}
    \ForEach{Active Filters $k \in \mathcal{K}_{t}$}{
      $\{\latent^{(k,n)}_{t-L+1:t}\}_{n} \leftarrow$ \textsc{Push$\left(\{\latent^{(k,n)}_{t}\}_{n}, \{\latent^{(k,n)}_{t-L:t-1}\}_{n}\right)$} \label{alg:push}\\
      \If{$j'_{k} > -1$}{ \label{alg:update}
        $\{\latent^{(k,n)}_{t-L+1:t}\}_{n} \leftarrow$ \textsc{ResampleAndSelect$\left(\{\latent^{(k,n)}_{t-L+1:t}\}_{n}, \obs^{(j'_{k})}_{t-L+1:t}\right)$} \label{alg:resample}\\
        $l_{k} \leftarrow \operatorname{Min}(l_{k} + 1, 30)$ \label{alg:increment}\\
      }
      \Else{
        $l_{k} \leftarrow l_{k} - 1$ \label{alg:decrement}\\
        \If{$l_{k} < 0$}{ \label{alg:deactivate1}
          \textsc{DeactivateFilter($k$)} \label{alg:deactivate2}
        }
      }
    }
  }
\caption{Proposed Multi-person Pose Tracking Method} \label{alg:pose_track}
\end{algorithm}

\section{Experiments} \label{sec:experiments}
\subsection{Comparison to State-of-the-art Method}
\begin{table*}[tb]
    \centering
    \caption{Multi-person Pose Tracking Performance on PoseTrack2018 Validation Dataset.}
    \label{tab:my_label}
    \begin{tabular}{cccccccc>{\columncolor[gray]{0.9}}c} \bhline{1pt}
         Method & \mota{Head} & \mota{Sho.} & \mota{Elb.} & \mota{Wri.} & \mota{Hip} & \mota{Knee} & \mota{Ank.} & \mota{Total} \\\hline
         MDPN-152-A~\cite{guo2018multi} & 50.9 & 55.5 & 65.0 & 49.0 & 48.7 & 50.5 & 45.1 & 50.6 \\
         Detect-and-Track~\cite{girdhar2018detect} & 61.7 & 65.5 & 57.3 & 45.7 & 54.3 & 53.1 & 45.7 & 55.2 \\
         Pose Flow~\cite{xiu2018pose} & 59.8 & 67.0 & 59.8 & 51.6 & 60.0 & 58.4 & 50.5 & 58.3 \\
         STAF~\cite{raaj2019efficient}  & - & - & - & - & - & - & - & 60.9 \\
         Simple-Baseline~\cite{xiao2018simple} & 73.9 & 75.9 & 63.7 & 56.1 &
 65.5 & 65.1 & 53.6 & 65.4 \\
         TML++~\cite{hwang2019pose} & 76.0 & 76.9 & 66.1 & 56.4 & 65.1 & 61.6 & 52.4 & 65.7 \\\hline
         Ours & 72.5 & 76.5 & 66.8 & 58.6 & 63.2 & 65.2 & 57.0 & \textbf{66.2} \\\bhline{1pt}
    \end{tabular}
\end{table*}
The main objective of this experiment was to demonstrate that the proposed method has advantages over the state-of-the-art pose tracking method~\cite{guo2018multi,girdhar2018detect,raaj2019efficient,xiao2018simple,hwang2019pose,xiu2018pose}.

Training and evaluation were conducted using the PoseTrack2018 dataset.
The annotations include 17 body keypoint locations and unique track IDs for multiple persons in the videos.
Training data $\mathcal{D}$ for were created from the training annotation data,
and the probabilistic neural pose predictor (Fig.~\ref{fig:predictor}) was trained using the Adam optimizer~\cite{kingma2014adam}.
Here, the learning rate was $10^{-3}$, and mini-batch size was 30.

We evaluated the performance of the proposed tracking method
using the validation data and official evaluation tool%
\footnote{\url{https://github.com/leonid-pishchulin/poseval}}.
Table \ref{tab:my_label} summarizes the MOTA scores obtained by the baseline and proposed method.
The best result obtained by the proposed method was a 66.2~MOTA score.
The proposed method outperformed both the winner of the ECCV’18 PoseTrack Challenge~\cite{xiao2018simple} (MOTA: 65.4) and the recent state-of-the-art method~\cite{raaj2019efficient} (MOTA: 65.7).
\subsection{Ablation Study}
\newcommand{\auncertainty}{\begin{tabular}{c}Aleatoric \\ Uncertainty \end{tabular}}
\newcommand{\euncertainty}{\begin{tabular}{c}Epistemic \\ Uncertainty \end{tabular}}%
\newcommand{\timeseqlen}{\begin{tabular}{c}Time-Seq. \\ Length $L$ \end{tabular}}%
\begin{table*}[tb]
    \centering
    \caption{Results of Ablation Study.} \label{tab:ablation}
    \label{tab:my_label}
    \begin{tabular}{ccccc>{\columncolor[gray]{0.9}}c} \bhline{1pt}
         Method & \auncertainty & \euncertainty & \timeseqlen & \mota{Total} & \texttt{num\_switches} \\\hline
         Ours & $\checkmark$ & $\checkmark$ & 15 & 66.1 & 232 \\
         Ours & $\checkmark$ & $\checkmark$ & 10 & 66.2 & \underline{\textbf{213}} \\
         Ours & $\checkmark$ & $\checkmark$ & 7 & 66.1 & 242 \\
         Ours & $\checkmark$ & $\checkmark$ & 3 & 66.0 & 266 \\\hline
         Ours & $\checkmark$ & & 10 & 66.1 & 252 \\
         Ours & & $\checkmark$ & 10 & 66.0 & 285 \\
         Ours & & & 10 & 65.9 & 303 \\\hline
         Simple-Baseline~\cite{xiao2018simple} & - & - & - & 65.4 & 407 \\\bhline{1pt}
    \end{tabular}
\end{table*}
This experiment was performed to clarify which component of the proposed method contributes to the overall improvement.
Here, variants of our method were prepared:
\textit{1)} both or either types of uncertainty were invalidated, and
\textit{2)} the length of time-sequence $L$ was varied.
Note that we removed epistemic uncertainty modeling by deactivating dropout in the fully-connected layer of the predictor,
and heteroscedastic aleatoric uncertainty was removed by fixing the value of $\preds$.

As a principle metric, we focused on \texttt{num\_switches} (an intermediate variable used to calculate MOTA) rather than MOTA in this experiment.
MOTA comprises of three variables: \texttt{num\_switches} as tracking error counts, and \texttt{num\_misses} and \texttt{num\_false\_positives} as pose estimation error counts (see the source code of the evaluation tool or \cite{bernardin2006multiple}).
We can distinctly compare the tracking performances by inputting shared pose estimation results to different tracking models and focusing on \texttt{num\_switches}.
Specifically, we focused on the total \texttt{num\_swithces} of the most frequently appeared keypoint (i.e., nose).

We also included an evaluation of the \textit{Simple-Baseline} method in this experiment to clearly demonstrate that
the above state-of-the-art result was achieved by the proposed tracking approach rather than other factors (e.g., more accurate pose estimation).
Since an evaluation of \texttt{num\_switches} was not conducted in a previous study~\cite{xiao2018simple},
we used our own implementation of the baseline method for this test.

\begin{figure*}[t]
  \centering
  \includegraphics[width=0.95\textwidth]{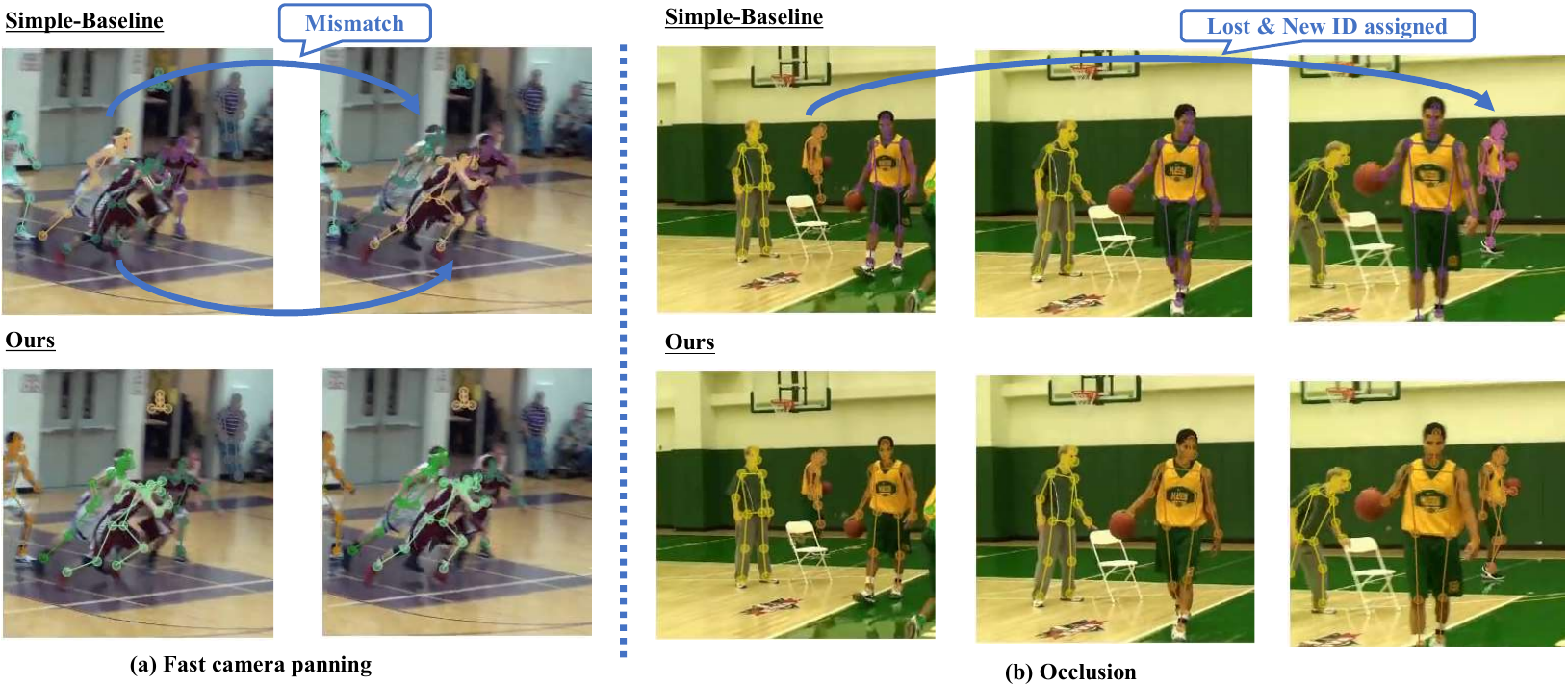}
  \vspace*{-3mm}
  \caption{
  Visualized results of \textit{Simple-Baseline} and proposed method.
  Estimated poses are color-coded according to track IDs.
  The baseline method fails to track some poses due to fast pose changes and occluions.
  In such difficult cases, the proposed method can track the poses successfully.
  }
  \label{fig:vis_ex}
\end{figure*}

The results of this ablation study are summarized in Table~\ref{tab:ablation},
which demonstrates that involving both types of uncertainty contributes to performance improvement.
Utilizing the time sequence input with LSTM is also effective.
By referring to longer pose contexts (e.g., $L=10,15$), the predictor can infer more plausibile hypotheses, which results in overall performance improvement.
However, parameter $L$ should be determined carefully because it affects computational complexity (i.e., the memory size and computational time of sequential LSTM forwarding)
and training stability.
A comparison of results obtained by \textit{Simple-Baseline} indicates that our best result achieves approximately 50\% of the baseline's tracking errors.
Fig.~\ref{fig:vis_ex} shows some visualized results obtained by the baseline and proposed method.

%

\section{Related Work} \label{sec:related_work}
Recently, uncertainty-aware modeling has been receiving increasing attention in robotics studies, including
reinforcement learning~\cite{chua2018deep,okada2019variational},
imitation learning~\cite{thakur2019uncertainty,silverio2019uncertainty}, 
motion and path planning~\cite{bowman2019intent,nardi2019uncertainty},
and unfamiliar situation detection~\cite{mcallister2019robustness}.
For example, Refs.~\cite{chua2018deep,okada2019variational} also incorporated the two types of uncertainty to forward dynamics modeling, thereby solving the major inherent problem of model-based reinforcement learning, i.e.,~the model-bias problem.
Approaches that are similar to our method have been proposed in previous tracking and SMC studies,
such as SMC based multi-object tracking~\cite{jaward2006multiple} and the combination of SMC and neural networks~\cite{gu2015neural,jonschkowski2018differentiable,karkus2018particle}.
However, to the best of our knowledge,
the proposed method is the first to apply SMC to multi-person pose tracking using a probabilistic neural network that incorporates the two types of uncertainty,
and the results provide a new state-of-the-art in the challenging and competitive benchmark task.


\section{Conclusion}
In this paper, we have proposed an SMC-based multi-person pose tracking method that utilizes a probabilistic neural pose predictor.
By incorporating epistemic uncertainty and heteroscedastic aleatoric uncertainty,
our pose predictor can propose diverse and plausible hypotheses for the next frame poses,
thereby solving the fragility of the sole hypothesis approach of the state-of-the-art baseline~\cite{xiao2018simple}.
In addition, recurrent neural modeling is introduced, which exploits long context information to make prediction robust against complex situations,
such as cases with significant occlusions.
The experimental results demonstrate that the proposed method achieves a state-of-the art MOTA score of 66.2 with approximately 50\% reduction in tracking errors from the baseline.

Other sophisticated uncertainty modeling could improve the proposed method's overall performance, such as
employing $\alpha$-divergence dropout~\cite{li2017dropout} and neural network ensembles~\cite{chua2018deep} for epistemic uncertainty modeling,
as well as replacing the output Gaussian distribution with a Gaussian Mixture Model by introducing Mixture Density Networks~\cite{gu2015neural,bishop1994mixture} for aleatoric uncertainty modeling.
In addition, inputting optical-flows to a neural predictor seems promising approach to utilize visual information,
and the end-to-end supervised learning of SMC~\cite{jonschkowski2018differentiable,karkus2018particle} is appealing to automatically design an effective likelihood function $p(\obs_{T}|\latent^{(n)}_{1:T},\obs_{1:T-1})$, which could improve the validity of the matching score compared to existing hand-crafted metrics, e.g.,~OKS .

The uncertainty- and context-aware concept for SMC proposed in this paper is simple, general, and strong, and we exepct that this concept is applicable to a variety of SMC-based robotics tasks, such as SLAM.
Other possible future work could include more challenging tracking tasks, such as 3D human pose tracking~\cite{h36m_pami} and dense human pose tracking~\cite{alp2018densepose}.

\bibliography{icra}
\bibliographystyle{ieeetr}

\end{document}